\documentclass{article}
\usepackage{ijcai24}

\usepackage{times}
\usepackage{soul}
\usepackage{url}
\usepackage[hidelinks]{hyperref}
\usepackage[utf8]{inputenc}
\usepackage[small]{caption}
\usepackage{graphicx}
\usepackage{amsmath}
\usepackage{amsthm}
\usepackage{booktabs}
\usepackage{algorithm}
\usepackage{algorithmic}
\usepackage[switch]{lineno}
\usepackage{amssymb}
\usepackage{subfig}
\usepackage{xcolor}
\usepackage{enumitem}
\usepackage{tikz}

\usepackage{blindtext}

\DeclareMathOperator*{\argmax}{arg\,max}
\DeclareMathOperator*{\argmin}{arg\,min}

\newcommand*\circled[1]{\tikz[baseline=(char.base)]{
            \node[shape=circle,draw,inner sep=2pt] (char) {#1};}}

\newtheorem{theorem}{Theorem}
\newtheorem{definition}{Definition}
\newtheorem{proposition}{Proposition}

\title{Offline Reinforcement Learning with Behavioral Supervisor Tuning}

\author{
Padmanaba Srinivasan
\and
William Knottenbelt
\affiliations
Imperial College London\\
\emails
\{ps3416, wjk\}@imperial.ac.uk
}

\begin{document}

\maketitle

\begin{abstract}
    Offline reinforcement learning (RL) algorithms are applied to learn performant, well-generalizing policies when provided with a static dataset of interactions. Many recent approaches to offline RL have seen substantial success, but with one key caveat: they demand substantial per-dataset hyperparameter tuning to achieve reported perf ormance, which requires policy rollouts in the environment to evaluate; this can rapidly become cumbersome. Furthermore, substantial tuning requirements can hamper the adoption of these algorithms in practical domains. In this paper, we present TD3 with \textbf{B}ehavioral \textbf{S}upervisor \textbf{T}uning (TD3-BST), an algorithm that trains an uncertainty model and uses it to guide the policy to select actions within the dataset support. TD3-BST can learn more effective policies from offline datasets compared to previous methods and achieves the best performance across challenging benchmarks without requiring per-dataset tuning. 
\end{abstract}

\section{Introduction}

Reinforcement learning (RL) is a method of learning where an agent interacts with an environment to collect experiences and seeks to maximize the reward provided by the environment. This typically follows a repeating cycle of experience collecting and improvement \cite{RN679}. This is termed \textit{online} RL due to the need for policy rollouts in the environment. Both on-policy and off-policy RL require some schedule of online interaction which, in some domains, can be infeasible due to experimental or environmental limitations \cite{RN839,RN840}. With such constraints, a dataset may instead be collected that consists of demonstrations by arbitrary (potentially multiple, unknown) behavior policies \cite{RN695} that may be suboptimal. Offline reinforcement learning algorithms are designed to recover optimal policies from such static datasets. 

\begin{figure}[h]
    \centering
    \includegraphics[width=1.0\columnwidth]{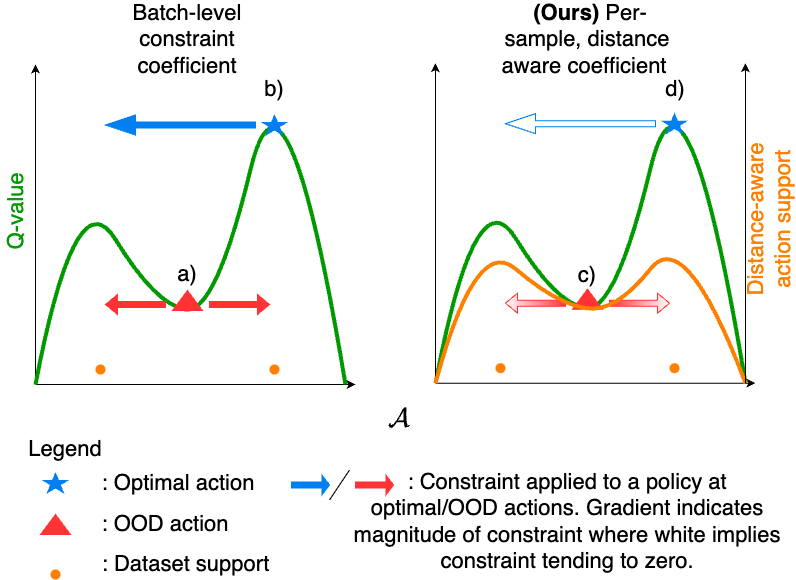}
    \caption{An illustration of our method versus typical, TD3-BC-like actor-constraint methods. \textbf{TD3-BC: } a) A policy selecting an OOD action is constrained to select in-dataset actions. b) A policy selecting the optimal action may be penalized for not selecting an in-dataset, but not in-batch, inferior action. \textbf{Our method: } c) A policy selecting OOD actions is drawn towards in-dataset actions with decreasing constraint coefficient as it moves closer to \textit{any} supported action. d) An optimal policy is not penalized for selecting an in-dataset action when the action is not contained in the current batch.}
\end{figure}

The primary challenge in offline RL is the evaluation of out-of-distribution (OOD) actions; offline datasets rarely offer support over the entire state-action space and neural networks overestimate values when extrapolating to OOD actions \cite{RN714,RN841,RN698,RN697}. If trained using standard off-policy methods, a policy will select any actions that maximize reward, which includes OOD actions. The difference between the rewards implied by the value function and the environment results in a distribution shift that can result in failure in real-world policy rollouts. Thus, offline RL algorithms must both maximize the reward and follow the behavioral policy, while having to potentially ``stitch'' together several suboptimal trajectories. The former requirement is usually satisfied by introducing a constraint on the actor to either penalize deviation from the behavior policy or epistemic uncertainty of the value function, or by regularizing the value function to directly minimize OOD action-values. 

Many recent approaches to offline RL \cite{RN842,RN843,RN844,RN847,RN851} demonstrate success in D4RL benchmarks \cite{RN731}, but demand the onerous task of per-dataset hyperparameter tuning \cite{RN852}. Algorithms that require substantial offline fine-tuning can be infeasible in real-world applications \cite{RN854}, hampering their adoption in favor of simpler, older algorithms \cite{RN855,RN856}. These older methods \cite{RN719,RN698,RN711} provide excellent ``bang-for-buck'' as their hyperparameters work well across a range of D4RL datasets.

\paragraph{Contributions}In this paper, we show how a trained uncertainty model can be incorporated into the regularized policy objective as a \textit{behavioral supervisor} to yield TD3 with behavioral supervisor tuning (TD3-BST). The key advantage of our method is the dynamic regularization weighting performed by the uncertainty network, which allows the learned policy to maximize $Q$-values around dataset modes. Evaluation on D4RL datasets demonstrates that TD3-BST achieves SOTA performance, and ablation experiments analyze the performance of the uncertainty model and the sensitivity of the parameters of the BST objective.

\section{Related Work}

Reinforcement learning is a framework for sequential decision making often formulated as a Markov decision process (MDP), $\mathcal{M} = \{\mathcal{S}, \mathcal{A}, R, p, p_0, \gamma\}$ with state space $\mathcal{S}$, action space $\mathcal{A}$, a scalar reward dependent on state and action $R(s, a)$, transition dynamics $p$, initial state distribution $p_0$ and discount factor $\gamma \in [0, 1)$ \cite{RN679}. RL aims to learn a policy $\pi \in \Pi$ that executes action $a = \pi(s)$ that will maximize the expected discounted reward ${J(\pi) = \mathbb{E}_{\tau \sim P_{\pi}(\tau)} \left[ \sum_{t=0}^{T} \gamma_t R(s_t, a_t) \right]}$ where ${P_{\pi}(\tau) = p_0 (s_0) \prod_{t=0}^{T} \pi(a_t \mid s_t) p(s_{t+1} \mid s_t, a_t)}$ is the trajectory under $\pi$. Rather than rolling out an entire trajectory, a state-action value function (Q function) is often used: ${Q_\pi (s, a) = \mathbb{E}_{\tau \sim P_{\pi}(\tau)} \left[ \sum_{t=0}^{T} \gamma_t r(s_t, a_t) \mid s_0 = s, a_0 = a \right]}$.

\subsection{Offline Reinforcement Learning}

Offline RL algorithms are presented with a static dataset $\mathcal{D}$ that consists of tuples $\{s, a, r, s'\}$ where $r \sim R(s, a)$ and $s' \sim p(\cdot \mid s, a)$. 

$\mathcal{D}$ has limited coverage over $\mathcal{S} \times \mathcal{A}$; hence, offline RL algorithms must constrain the policy to select actions within the dataset support. To this end, algorithms employ one of three approaches: 1) policy constraints; 2) critic regularization; or 3) uncertainty penalization. 

\paragraph{Policy constraint}Policy constraints modify the actor's objective only to minimize divergence from the behavior policy. Most simply, this adds a constraint term \cite{RN719,RN842} to the policy objective:
\begin{equation}
    \label{eq: policy constraint objective}
    \argmax_\pi \mathbb{E}_{\{s, a\} \sim \mathcal{D}} \left[ Q(s, \pi(s)) - \alpha D(\pi, \pi_{\beta}) \right],
\end{equation}

\noindent where $\alpha$ is a scalar controlling the strength of regularization, $D(\cdot, \cdot)$ is a divergence function between the policy $\pi$ and the behavior policy $\pi_{\beta}$. In offline RL, we do not have access to $\pi_{\beta}$; some prior methods attempt to estimate it empirically \cite{RN703,RN876} which is challenging when the dataset is generated by a mixture of policies. Furthermore, selecting the constraint strength can be challenging and difficult to generalize across datasets with similar environments \cite{RN842,RN703}. 

Other policy constraint approaches use weighted BC \cite{RN706,RN711,RN851} or (surrogate) BC constraints \cite{RN844,RN699,RN876}. The former methods may be too restrictive as they do not allow OOD action selection, which is crucial to improve performance \cite{RN813}. The latter methods may still require substantial tuning and in addition to training if using model-based score methods. Other methods impose architectural constraints \cite{RN697,RN684} that parameterize separate BC and reward-maximizing policy models.

\paragraph{Critic Regularization}Critic regularization methods directly address the OOD action-value overestimation problem by penalizing large values for adversarially sampled actions \cite{RN703}. 

\paragraph{Ensembles}Employing an ensemble of neural network estimators is a commonly used technique for prediction with a measure of epistemic uncertainty \cite{RN860}. A family of offline RL methods
employ large ensembles of value functions \cite{RN770} and make use of the diversity of randomly initialized ensembles to implicitly reduce the selection of OOD actions or directly penalize the variance of the reward in the ensemble \cite{RN771,RN679}. 

\paragraph{Model-Based Uncertainty Estimation}Learning an uncertainty model of the dataset is often devised analogously to exploration-encouraging methods used in online RL, but, employing these for anti-exploration instead \cite{RN723}. An example is SAC-RND which directly adopts such an approach \cite{RN847}. Other algorithms include DOGE \cite{RN844} which trains a model to estimate uncertainty as a distance to dataset action and DARL \cite{RN873} which uses distance to random projections of state-action pairs as an uncertainty measure. As a whole, these methods optimize a distance $d(\cdot, \cdot) \geq 0$ that represents the uncertainty of an action.

\subsection{Uncertainty Estimation}

Neural networks are known to predict confidently even when presented with OOD samples \cite{RN863,RN864,RN868}. A classical approach to OOD detection is to fit a generative model to the dataset that produces a high probability for in-dataset samples and a low probability for OOD ones. These methods work well for simple, unimodal data but can become computationally demanding for more complex data with multiple modes. Another approach trains classifiers that are leveraged to become finer-grained OOD detectors \cite{RN869}. In this work, we focus on Morse neural networks \cite{RN870}, an approach that trains a generative model to produce an unnormalized density that takes on value 1 at the dataset modes.

\section{Preliminaries}

A Morse neural network produces an unnormalized density $M (x) \in [0, 1]$ on an embedding space $\mathbb{R}^e$ \cite{RN870}. A Morse network can produce a density in $\mathbb{R}^{e}$ that attains a value of 1 at mode submanifolds and decreases towards 0 when moving away from the mode. The rate at which the value decreases is controlled by a Morse Kernel.

\begin{definition}[Morse Kernel]
    \label{def: morse kernel}
    A Morse Kernel is a positive definite kernel $K$. When applied in a space $Z = \mathbb{R}^k$, the kernel $K(z_1, z_2)$ takes values in the interval $[0, 1]$ where $K(z_1, z_2) = 1$ iff $z_1 = z_2$.
\end{definition}

All kernels of the form $K(z_1, z_2) = e^{-D(z_1, z_2)}$ where $D(\cdot,\cdot)$ is a divergence \cite{RN874} are Morse Kernels. Examples include common kernels such as the Radial Basis Function (RBF) Kernel,  
\begin{equation}
    K_{RBF} (z_1, z_2) = e^{- \frac{\lambda^2}{2} \mid\mid z_1 - z_2 \mid\mid^2}.
\end{equation}

The RBF kernel and its derivatives decay exponentially, leading learning signals to vanish rapidly. An alternative is the ubiquitous Rational Quadratic (RQ) kernel:

\begin{equation}
    K_{RQ} (z_1, z_2) = \left(1 + \frac{\lambda ^ 2}{2 \kappa} \mid\mid z_1 - z_2 \mid\mid^2 \right)^{-\kappa}
\end{equation}

\noindent where $\lambda$ is a scale parameter in each kernel. The RQ kernel is a scaled mixture of RBF kernels controlled by $\kappa$ and, for small $\kappa$, decays much more slowly \cite{RN887}. 

Consider a neural network that maps from a feature space into a latent space $f_{\phi}: X \rightarrow Z$, with parameters $\phi$, $X \in \mathbb{R}^d$ and $Z \in \mathbb{R}^k$. A Morse Kernel can impose structure on the latent space. 

\begin{definition}[Morse Neural Network]
    \label{def: morse neural network}
    A Morse neural network is a function $f_{\phi}: X \rightarrow Z$ in combination with a Morse Kernel on $K(z, t)$ where $t \subset Z$ is a target, chosen as a hyperparameter of the model. The Morse neural network is defined as $M_{\phi} (x) = K(f_{\phi} (x), t)$.
\end{definition}

Using Definition~\ref{def: morse kernel} we see that $M_{\phi}(x) \in [0, 1]$, and when $M_{\phi}(x) = 1$, $x$ corresponds to a mode that coincides with the level set of the submanifold of the Morse neural network. Furthermore, $M_{\phi}(x)$ corresponds to the \textit{certainty} of the sample $x$ being from the training dataset, so $1 - M_{\phi}(x)$ is a measure of the epistemic uncertainty of $x$.

The function $-\log M_{\phi} (x)$ measures a squared distance, $d(\cdot, \cdot)$, between $f_{\phi} (x)$ and the closest mode in the latent space at $m$:
\begin{equation}
    \label{eq: closest neighbor property}
    d(z) = \min_{m \in M} d(z, m),
\end{equation}

\noindent where $M$ is the set of all modes. This encodes information about the topology of the submanifold and satisfies the Morse--Bott non-degeneracy condition \cite{RN872}.

The Morse neural network offers the following properties:

\begin{enumerate}[label=\protect\circled{\arabic*}]
    \label{list: morse network properties}
    \item $M_{\phi}(x) \in [0, 1]$.
    \item $M_{\phi}(x) = 1$ at its mode submanifolds.
    \item $-\log M_{\phi} (x) \geq 0$ is a squared distance that satisfies the Morse--Bott non-degeneracy condition on the mode submanifolds. 
    \item As $M_{\phi}(x)$ is an exponentiated squared distance, the function is also distance aware in the sense that as $f_{\phi} (x) \rightarrow t, M_{\phi}(x) \rightarrow 1$.
\end{enumerate}

Proof of each property is provided in the appendix.

\section{Policy Constraint with a Behavioral Supervisor}

We now describe the constituent components of our algorithm, building on the Morse network and showing how it can be incorporated into a policy-regularized objective. 

\subsection{Morse Networks for Offline RL}

The target $t$ is a hyperparameter that must be chosen. Experiments in \cite{RN870} use simple, toy datasets with classification problems that perform well for categorical $t$. We find that using a static label for the Morse network yields poor performance; rather than a labeling model, we treat $f_{\phi}$ as a perturbation model that produces an action $f_{\phi} (s, a) = \hat{a}$ such that $\hat{a} = a$ if and only if $s, a \sim \mathcal{D}$.

An offline RL dataset $\mathcal{D}$ consists of tuples $\{s, a, r, s'\}$ where we assume $\{s, a\}$ pairs are i.i.d.\ sampled from an unknown distribution. The Morse network must be fitted on $N$ state-action pairs $[\{s_1, a_1,\}, ..., \{s_N, a_N\}]$ such that $M_{\phi} (s_i, a_j) = 1, \forall i, j \in 1,...,N$ ] only when $i = j$. 

We fit a Morse neural network to minimize the KL divergence between unnormalized measures \cite{RN874} following \cite{RN870}, $D_{\text{KL}} (\mathcal{D}(s, a) \mid\mid M_{\phi} (s, a))$:
\begin{equation}
    \min_{\phi} \mathbb{E}_{s, a \sim \mathcal{D}} \left[ \log \frac{\mathcal{D} (s, a)}{M_{\phi} (s, a)} \right] + \int M_{\phi} (s, a) - \mathcal{D} (s, a)\, da.
\end{equation}

With respect to $\phi$, this amounts to minimizing the empirical loss: 
\begin{align}
    \label{eq: morse training objective}
    \nonumber L(\phi) = -&\frac{1}{N} \sum_{s, a \sim \mathcal{D}} \log K( f_{\phi} (s, a), a) \\ + &\frac{1}{N} \sum_{\substack{s \sim \mathcal{D} \\ a_{\bar{\mathcal{D}}} \sim \mathcal{D}_{\text{uni}}}} K( f_{\phi} (s, a_{\text{u}}), a_{\text{u}}),
\end{align}
\noindent where $a_{\text{u}}$ is an action sampled from a uniform distribution over the action space $\mathcal{D}_{\text{uni}}$. 

A learned Morse density is well suited to modeling ensemble policies \cite{RN880}, more flexibly \cite{RN870,RN703,RN876} and without down-weighting good, in-support actions that have low density under the behavior policy \cite{RN754} as {\em all} modes have unnormalized density value 1.

A Morse neural network can be expressed as an energy-based model (EBM) \cite{RN884}:
\begin{proposition}
    A Morse neural network can be expressed as an energy-based model: $E_{\phi} (x) = e^{-\log M_{\phi} (x)}$ where $M_{\phi}: \mathbb{R}^{d} \rightarrow \mathbb{R}$. 
\end{proposition}

Note that the EBM $E_{\phi}$ is itself unnormalized. Representing the Morse network as an EBM allows analysis analogous to \cite{RN885}. 

\begin{theorem}
    \label{thm: ebm bounded error}
    For a set-valued function $F(x): x \in \mathbb{R}^{m} \rightarrow \mathbb{R}^{n} \text{\textbackslash}\{\emptyset\}$, there exists a continuous function $g: \mathbb{R}^{m + n} \rightarrow \mathbb{R}$ that is approximated by a continuous function approximator $g_{\phi}$ with arbitrarily small bounded error $\epsilon$. This ensures that any point on the graph $F_{\phi} (x) = \argmin_{y} g_{\phi} (x, y)$ is within distance $\epsilon$ of $F$. 
\end{theorem}

We refer the reader to \cite{RN885} for a detailed proof.

The theorem assumes that $F(x)$ is an implicit function and states that the error at the level-set (i.e.\ the modes) of $F(x)$ is small.

\subsection{TD3-BST}

We can use the Morse network to design a regularized policy objective. Recall that policy regularization consists of $Q$-value maximization and minimization of a distance to the behavior policy (Equation~\ref{eq: policy constraint objective}). We reconsider the policy regularization term and train a policy that minimizes uncertainty while selecting actions close to the behavior policy. Let $C^{\pi}(s, a)$ denote a measure of uncertainty of the policy action. We solve the following optimization problem:
\begin{align}
    \pi^{i+1} = \argmin_{\pi \in \Pi} \mathbb{E}_{a \sim \pi (\cdot \mid s)} \left[ C^{\pi}(s, a) \right]
    \\
    \text{s.t.}\quad D_{\text{KL}} \left( \pi(\cdot \mid s) \mid\mid \pi_{\beta} (\cdot \mid s) \right) \leq \epsilon.
\end{align}

This optimization problem requires an explicit behavior model, which is difficult to estimate and using an estimated model has historically returned mixed results \cite{RN697,RN684}. Furthermore, this requires direct optimization through $C^{\pi}$ which may be subject to exploitation. Instead, we enforce this \textit{implicitly} by deriving the solution to the constrained optimization to obtain a closed-form solution for the actor \cite{RN705,RN706}. Enforcing the KKT conditions we obtain the Lagrangian:
\begin{equation}
    L(\pi, \mu) = \mathbb{E}_{a \sim \pi(\cdot \mid s)} \left[ C^{\pi} (s, a) \right] + \mu (\epsilon - D_{\text{KL}} (\pi \mid\mid \pi_{\beta})).
\end{equation}

Computing $\frac{\partial L}{\partial \pi}$ and solving for $\pi$ yields the uncertainty minimizing solution ${\pi^{C^*}(a \mid s) \propto \pi_{\beta}(a \mid s) e^{\frac{1}{\mu} C^{\pi}(s, a)}}$. When learning the parametric policy $\pi_{\psi}$, we project the non-parametric solution into the policy space as a (reverse) KL divergence minimization of $\pi_{\psi}$ under the data distribution $\mathcal{D}$: 
\begin{align}
     \argmin_{\psi} \mathbb{E}_{s \sim \mathcal{D}} \left[ D_{\text{KL}} \left( \pi^{C^*}(\cdot \mid s) \mid\mid \pi_{\psi}(\cdot \mid s) \right) \right]
    \\
    = \argmin_{\psi} \mathbb{E}_{s \sim \mathcal{D}} \left[ D_{\text{KL}} \left( \pi_{\beta}(a \mid s) e^{\frac{1}{\mu} C^{\pi}(s, a)} \mid\mid \pi_{\psi}(\cdot \mid s) \right) \right]
    \\
    = \argmin_{\psi} \mathbb{E}_{s, a \sim \mathcal{D}} \left[ -\log \pi_{\psi} (a \mid s) e^{\frac{1}{\mu} C^{\pi}(s, a)} \right],
\end{align}

\noindent which is a weighted maximum likelihood update where the supervised target is sampled from the dataset $\mathcal{D}$ and ${C^{\pi}(s, a) = 1 - M_{\phi} (s, \pi_{\psi} (s))}$. This avoids explicitly modeling the behavior policy and uses the Morse network uncertainty as a \textit{behavior supervisor} to dynamically adjust the strength of behavioral cloning. We provide a more detailed derivation in the appendix.

\paragraph{Interpretation}\quad Our regularization method shares similarities with other weighted regression algorithms \cite{RN706,RN705,RN711} which weight the advantage of an action compared to the dataset/replay buffer action. Our weighting can be thought of as a measure of \textit{disadvantage} of a policy action in the sense of how OOD it is.

We make modifications to the behavioral cloning objective. From Morse network property \circled{1} we know $M_{\phi} \in [0, 1]$, hence ${1 \leq e^{\frac{1}{\mu} C^{\pi}} \leq e^{\frac{1}{\mu}}}$, i.e.\ the lowest possible disadvantage coefficient is 1. To minimize the coefficient in the mode, we require it to approach 0 when near a mode. We adjust the weighted behavioral cloning term and add $Q$-value maximization to yield the regularized policy update:
\begin{align}
    \label{eq: bst policy update}
    \nonumber \pi^{i+1} \leftarrow \argmax_{\pi} \mathbb{E}_{\substack{s, a \sim \mathcal{D}, \\ a_{\pi} \sim \pi^{i} (s)}} [ \frac{1}{Z_Q} Q^{i+1} (s, a_{\pi}) \\ - (e^{\frac{1}{\mu}C^{\pi}(s, a)} - 1) ( a_{\pi} - a )^2 ],
\end{align}

\noindent where $\mu$ is the Lagrangian multiplier that controls the magnitude of the disadvantage weight and ${Z_Q = \frac{1}{N} \sum_{n = 1}^{N} \lvert Q(s, a_{\pi}) \rvert}$ is a scaling term detached from the gradient update process \cite{RN719},  necessary as $Q(s, a)$ can be arbitrarily large and the BC-coefficient is upper-bounded at $e^{\frac{1}{\mu}}$. 

The value function update is given by:
\begin{align}
    \label{eq: bst value update}
    Q^{i+1} \leftarrow \argmin_{Q} \mathbb{E}_{s, a, s' \sim \mathcal{D}} [ (y - Q^{i} (s, a))^2 ],
\end{align}

\noindent with $y = r(s,a) + \gamma \mathbb{E}_{s' \sim \bar{\pi}(s')} \bar{Q} (s', a')$ where $\bar{Q}$ and $\bar{\pi}$ are target value and policy functions, respectively.

\subsection{Controlling the Tradeoff Constraint}

\begin{figure*}[ht]
    \centering
    \subfloat[\centering $\lambda = 0.1$]{{\includegraphics[width=0.15\textwidth]{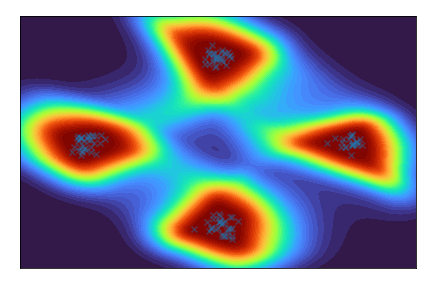} }}%
    \subfloat[\centering $\lambda = 0.5$]{{\includegraphics[width=0.15\textwidth]{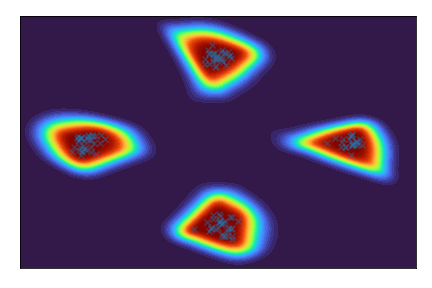} }}%
    \subfloat[\centering $\lambda = 1.0$]{{\includegraphics[width=0.15\textwidth]{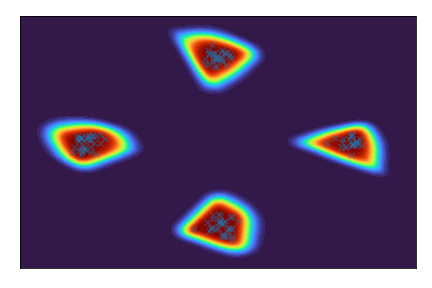} }}%
    \subfloat[\centering $\lambda = 2.0$]{{\includegraphics[width=0.15\textwidth]{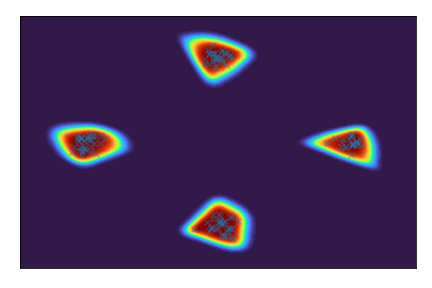} }}%
    \subfloat[\centering Ground Truth]{{\includegraphics[width=0.15\textwidth]{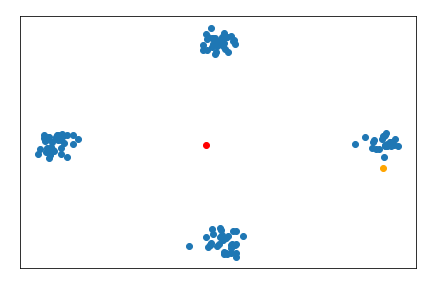} }}%
    \subfloat[\centering Density $\lambda = 1.0$]{{\includegraphics[width=0.15\textwidth]{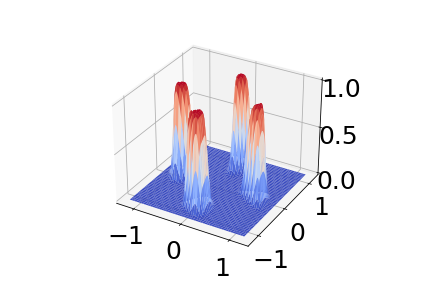} }}%
    \caption{a-d: Contour plots of unnormalized densities $(\in [0, 1])$ produced by a Morse network for increasing $\lambda$ with ground truth actions included as \textcolor{blue}{$\times$} marks. e: Ground truth actions (\textcolor{blue}{blue}) in the synthetic dataset, the MLE action (\textcolor{red}{red}). A Morse certainty weighted MLE model can select actions in a single mode, in this case, the (right-hand side) mode centred at [0.8, 0.0] (\textcolor{orange}{orange}). Weighting a divergence constraint using a Morse (un)certainty will encourage the policy to select actions near the modes of $M_{\phi}$ that maximize reward.}
    \label{fig: lambda illustrations}
\end{figure*}

Tuning TD3-BST is straightforward; the primary hyperparameters of the Morse network consist of the choice and scale of the kernel, and the temperature $\mu$. Increasing $\lambda$ for higher dimensional actions ensures that the high certainty region around modes remains tight. Prior empirical work has demonstrated the importance of allowing some degree of OOD actions \cite{RN770}; in the TD3-BST framework, this is dependent on $\lambda$. In Figure~\ref{fig: lambda illustrations} we provide a didactic example of the effect of $\lambda$. We construct a dataset consisting of 2-dimensional actions in $[-1, 1]$ with means at the four locations $\{[0.0, 0.8], [0.0, -0.8], [0.8, 0.0], [-0.8, 0.0]\}$ and each with standard deviation $0.05$. We sample $M = 128$ points, train a Morse network and plot the density produced by the Morse network for $\lambda = \{\frac{1}{10}, \frac{1}{2}, 1.0, 2.0\}$. A behavioral cloning policy learned using vanilla MLE where all targets are weighted equally results in an OOD action being selected. Training using Morse-weighted BC downweights the behavioral cloning loss for far away modes, enabling the policy to select and minimize error to a single mode.

\subsection{Algorithm Summary}

\paragraph{Fitting the Morse Network}The TD3-BST training procedure is described in Algorithm~\ref{algo: td3-bst training pseudocode}. The first phase fits the Morse network for $T_{M}$ gradient steps.

\paragraph{Actor--Critic Training}In the second phase of training, a modified TD3-BC procedure is used for $T_{\text{AC}}$ iterations with alterations highlighted in \textcolor{red}{red}. 

We provide full hyperparameter details in the appendix.

\begin{algorithm}[tb]
    \caption{TD3-BST Training Procedure Outline. The policy is updated once for every $m=2$ critic updates, as is the default in TD3.}
    \label{algo: td3-bst training pseudocode}
    \textbf{Input}: Dataset $\mathcal{D} = \{s, a, r, s'\}$\\
    \textbf{Initialize}: Initialize Morse network $M_{\phi}$. \\
    \textbf{Output}: Trained Morse network $M_{\phi}$. \\
    \begin{algorithmic}[0] 
        \STATE Let $t=0$.
        \FOR{{t = 1} \TO {$T_M$}}
        \STATE Sample minibatch $(s, a) \sim \mathcal{D}$
        \STATE Sample random actions $a_{\bar{\mathcal{D}}} \sim \mathcal{D}_{\text{uni}}$ for each state $s$
        \STATE Update $\phi$ by minimizing Equation~\ref{eq: morse training objective}
        \ENDFOR
    \end{algorithmic}
    \textbf{Initialize}: Initialize policy network $\pi_{\psi}$, critic $Q_{\theta}$, target policy $\bar{\psi} \leftarrow \psi$ and target critic $\bar{\theta} \leftarrow \theta$. \\
    \textbf{Output}: Trained policy $\pi$. \\
    \begin{algorithmic}[0] 
        \STATE Let $t=0$.
        \FOR{{t = 1} \TO {$T_{AC}$}}
        \STATE Sample minibatch $(s, a, r, s') \sim \mathcal{D}$
        \STATE Update $\theta$ using Equation~\ref{eq: bst value update}
        \IF {$t\ \text{mod}\ m = 0$}
            \STATE Obtain $a_{\pi} = \pi(s)$
            \color{red}
            \STATE Update $\psi$ using Equation~\ref{eq: bst policy update}
            \color{black}
            \STATE Update target networks $\bar{\theta} \leftarrow \rho \theta + (1 - \rho) \bar{\theta}$, $\bar{\psi} \leftarrow \rho \psi + (1 - \rho) \bar{\psi}$
        \ENDIF
        \ENDFOR
    \STATE \textbf{return} $\pi$
    \end{algorithmic}
\end{algorithm}

\section{Experiments}

In this section, we conduct experiments that aim to answer the following questions:
\begin{itemize}
    \item How does TD3-BST compare to other baselines, with a focus on comparing to newer baselines that use per-dataset tuning?
    \item Can the BST objective improve performance when used with one-step methods (IQL) that perform in-sample policy evaluation?
    \item How well does the Morse network learn to discriminate between in-dataset and OOD actions?
    \item How does changing the kernel scale parameter $\lambda$ affect performance? 
    \item Does using independent ensembles, a second method of uncertainty estimation, improve performance?
\end{itemize}

We evaluate our algorithm on the D4RL benchmark \cite{RN731}, including the Gym Locomotion and challenging Antmaze navigation tasks.

\subsection{Comparison with SOTA Methods}

We evaluate TD3-BST against the older, well known baselines of TD3-BC \cite{RN719}, CQL \cite{RN698}, and IQL \cite{RN711}. There are more recent methods that consistently outperform these baselines; of these, we include SQL \cite{RN851}, SAC-RND \cite{RN847}, DOGE \cite{RN844}, VMG \cite{RN843}, ReBRAC \cite{RN842}, CFPI \cite{RN876} and MSG \cite{RN771} (to our knowledge, the best-performing ensemble-based method). It is interesting to note that most of these baselines implement policy constraints, except for VMG (graph-based planning) and MSG (policy constraint using a large, independent ensemble). We note that {\em all} the aforementioned SOTA methods (except SQL) report scores with per-dataset tuned parameters in stark contrast with the older TD3-BC, CQL, and IQL algorithms, which use the same set of hyperparameters in each D4RL domain. All scores are reported with 10 evaluations in Locomotion and 100 in Antmaze across five seeds.

We present scores for D4RL Gym Locomotion in Table~\ref{tab: gym locomotion results}. TD3-BST achieves best or near-best results compared to all previous methods and recovers expert performance on five of nine datasets. The best performing prior methods include SAC-RND and ReBRAC, both of which require per-dataset tuning of BRAC-variant algorithms \cite{RN699}. 

We evaluate TD3-BST on the more challenging Antmaze tasks which contain a high degree of suboptimal trajectories and follow a sparse reward scheme that requires algorithms to \textit{stitch} together several trajectories to perform well. TD3-BST achieves the best scores overall in Table~\ref{tab: antmaze results}, especially as the maze becomes more complex. VMG and MSG are the best-performing prior baselines and TD3-BST is far simpler and more efficient in its design as a variant of TD3-BC. The authors of VMG report the best scores from checkpoints rather than from the final policy. MSG report scores from ensembles with both 4 and 64 critics of which the best scores included here are from the 64-critic variant.

We pay close attention to SAC-RND, which, among all baselines, is most similar in its inception to TD3-BST. SAC-RND uses a random and trained network pair to produce a dataset-constraining penalty. SAC-RND achieves consistent SOTA scores on locomotion datasets, but fails to deliver commensurate performance on Antmaze tasks. TD3-BST performs similarly to SAC-RND in locomotion and achieves SOTA scores in Antmaze. 

\begin{table*}
    \centering
    \resizebox{0.9\textwidth}{!}{
    \begin{tabular}{lrrrr|rrrr|r}
        \toprule
        Dataset  & TD3-BC & CQL & IQL & SQL & $\text{SAC-RND}^{1}$ & DOGE & ReBRAC & CFPI & \textbf{TD3-BST (ours)} \\
        \midrule
        \texttt{halfcheetah-m} & 48.3 & 44.0 & 47.4 & 48.3 & \textbf{66.6} & 45.3 & \underline{65.6} & 52.1 & 62.1 $\pm$ 0.8 \\
        \texttt{hopper-m} & 59.3 & 58.5 & 66.3 & 75.5 & 97.8 & 98.6 & \underline{102.0} & 86.8 & \textbf{102.9 $\pm$ 1.3} \\
        \texttt{walker2d-m} & 83.7 & 72.5 & 78.3 & 84.2 & \textbf{91.6} & 86.8 & 82.5 & 88.3 & \underline{90.7 $\pm$ 2.5} \\
        \texttt{halfcheetah-m-r} & 44.6 & 45.5 & 44.2 & 44.8 & 42.8 & \textbf{54.9} & 51.0 & 44.5 & \underline{53.0 $\pm$ 0.7} \\
        \texttt{hopper-m-r} & 60.9 & 95.0 & 94.7 & 99.7 & \underline{100.5} & 76.2 & 98.1 & 93.6 & \textbf{101.2 $\pm$ 4.9} \\
        \texttt{walker2d-m-r} & 81.8 & 77.2 & 73.9 & 81.2 & 88.7 & \underline{87.3} & 77.3 & 78.2 & \textbf{90.4 $\pm$ 8.3} \\
        \texttt{halfcheetah-m-e} & 90.7 & 91.6 & 86.7 & 94.0 & \textbf{107.6} & 78.7 & \underline{101.1} & 97.3 & 100.7 $\pm$ 1.1 \\
        \texttt{hopper-m-e} & 98.0 & 105.4 & 91.5 & \textbf{111.8} & 109.8 & 102.7 & 107.0 & 104.2 & \underline{110.3 $\pm$ 0.9} \\
        \texttt{walker2d-m-e} & 110.1 & 108.8 & 109.6 & 110.0 & 105.0 & 110.4 & \underline{111.6} & \textbf{111.9} & 109.4 $\pm$ 0.2 \\
        \bottomrule
    \end{tabular}
    }
    \caption{Normalized scores on D4RL Gym Locomotion datasets. VMG scores are excluded because this method performs poorly and the authors of MSG do not report numerical results on locomotion tasks. Prior methods are grouped by those that do not perform per-dataset tuning and those that do. $^{1}$ SAC-RND in addition to per-dataset tuning, is trained for 3 million gradient steps. Though not included here, ensemble methods may perform better than the best non-ensemble methods on some datasets, albeit still requiring per-dataset tuning to achieve their reported performance. Top scores are in \textbf{bold} and second-best are \underline{underlined}.}
    \label{tab: gym locomotion results}
\end{table*}

\begin{table*}
    \centering
    \resizebox{0.9\textwidth}{!}{
    \begin{tabular}{lrrrr|rrrrrr|r}
        \toprule
        Dataset  & TD3-BC & CQL & IQL & SQL & $\text{SAC-RND}^{1}$ & DOGE & $\text{VMG}^{2}$ & ReBRAC & CFPI & $\text{MSG}^{3}$ & \textbf{TD3-BST (ours)} \\
        \midrule
        \texttt{-umaze} & 78.6 & 74.0 & 87.5 & 92.2 & 97.0 & 97.0 & 93.7 & \underline{97.8} & 90.2 & \textbf{98.6} & \underline{97.8 $\pm$ 1.0} \\
        \texttt{-umaze-d} & 71.4 & 84.0 & 62.2 & 74.0 & 66.0 & 63.5 & \textbf{94.0} & 88.3 & 58.6 & 81.8 & \underline{91.7 $\pm$ 3.2} \\
        \texttt{-medium-p} & 10.6 & 61.2 & 71.2 & 80.2 & 74.7 & 80.6 & 82.7 & 84.0 & 75.2 & \underline{89.6} & \textbf{90.2 $\pm$ 1.8} \\
        \texttt{-medium-d} & 3.0 & 53.7 & 70.0 & 79.1 & 74.7 & 77.6 & 84.3 & 76.3 & 72.2 & \underline{88.6} & \textbf{92.0 $\pm$ 3.8} \\
        \texttt{-large-p} & 0.2 & 15.8 & 39.6 & 53.2 & 43.9 & 48.2 & 67.3 & 60.4 & 51.4 & \underline{72.6} & \textbf{79.7 $\pm$ 7.6} \\
        \texttt{-large-d} & 0.0 & 14.9 & 47.5 & 52.3 & 45.7 & 36.4 & \underline{74.3} & 54.4 & 52.4 & 71.4 & \textbf{76.1 $\pm$ 4.7} \\
        \bottomrule
    \end{tabular}
    }
    \caption{Normalized scores on D4RL Antmaze datasets. $^{1}$ SAC-RND is trained for three million gradient steps. $^{2}$ VMG reports scores from the best-performing checkpoint rather than from the final policy; despite this, TD3-BST still outperforms VMG in all datasets except \texttt{-umaze-diverse}. $^{3}$ for MSG we report the best score among the reported scores of all configurations, also, MSG is trained for two million steps. Prior methods are grouped by those that do not perform per-dataset tuning and those that do. Other ensemble-based methods are not included, as MSG achieves higher performance. Top scores are in \textbf{bold} and second-best are \underline{underlined}.}
    \label{tab: antmaze results}
\end{table*}

\subsection{Improving One-Step Methods}

One-step algorithms learn a policy from an offline dataset, thus remaining on-policy \cite{RN886,RN679}, and using weighted behavioral cloning \cite{RN882,RN711}. Empirical evaluation by \cite{RN813} suggests that advantage-weighted BC is too restrictive and relaxing the policy objective to Equation~\ref{eq: policy constraint objective} can lead to performance improvement. We use the BST objective as a drop-in replacement for the policy improvement step in IQL \cite{RN711} to learn an optimal policy while retaining in-sample policy evaluation.

We reproduce IQL results and report scores for IQL-BST, both times using a deterministic policy \cite{RN820} and identical hyperparameters to the original work in Table~\ref{tab: IQL, IQL-BST antmaze results}. Reproduced IQL closely matches the original results, with slight performance reductions on the \texttt{-large} datasets. Relaxing weighted-BC with a BST objective leads to improvements in performance, especially on the more difficult \texttt{-medium} and \texttt{-large} datasets. To isolate the effect of the BST objective, we do not perform any additional tuning.

\begin{table}
    \centering
    \resizebox{0.80\columnwidth}{!}{
    \begin{tabular}{lrr}
        \toprule
        Dataset  & IQL (reproduced) & IQL-BST \\
        \midrule
        \texttt{-umaze} & 87.6 $\pm$ 4.6 & \textbf{90.8 $\pm$ 2.1} \\
        \texttt{-umaze-d} & \textbf{64.0 $\pm$ 5.2} & 63.1 $\pm$ 3.7 \\
        \texttt{-medium-p} & 70.7 $\pm$ 4.3 & \textbf{80.3 $\pm$ 1.3} \\
        \texttt{-medium-d} & 73.8 $\pm$ 5.9 & \textbf{84.7 $\pm$ 2.0} \\
        \texttt{-large-p} & 35.2 $\pm$ 8.4 & \textbf{55.4 $\pm$ 3.2} \\
        \texttt{-large-d} & 40.7 $\pm$ 9.2 & \textbf{51.6 $\pm$ 2.6} \\
        \bottomrule
    \end{tabular}
    }
    \caption{Normalized scores on D4RL Antmaze datasets for IQL and IQL-BST. We use hyperparameters identical to the original IQL paper and use Equation~\ref{eq: bst policy update} as the policy objective.}
    \label{tab: IQL, IQL-BST antmaze results}
\end{table}

\subsection{Ablation Experiments}

\begin{figure}[h]
    \centering
    \subfloat{{\includegraphics[width=0.48\columnwidth]{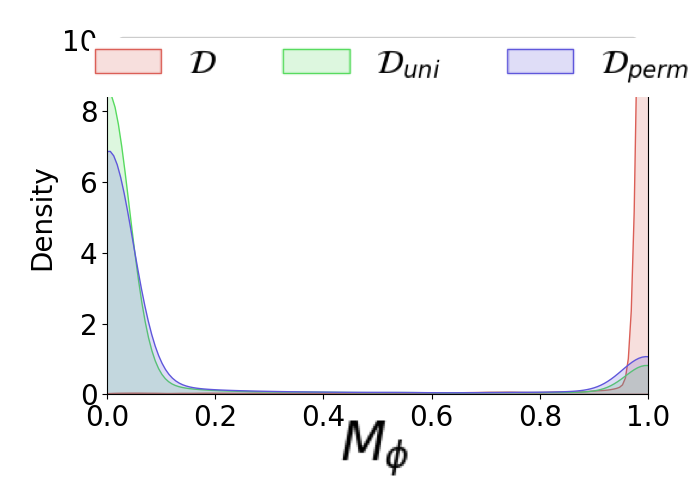} }}%
    \subfloat{{\includegraphics[width=0.48\columnwidth]{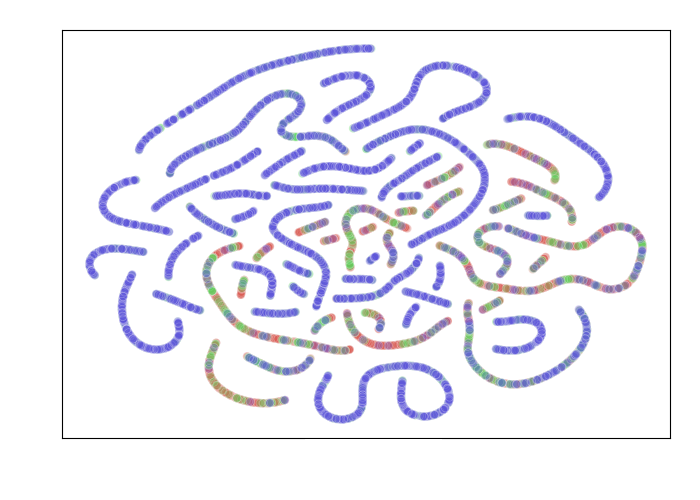} }}%
    \\
    \subfloat{{\includegraphics[width=0.48\columnwidth]{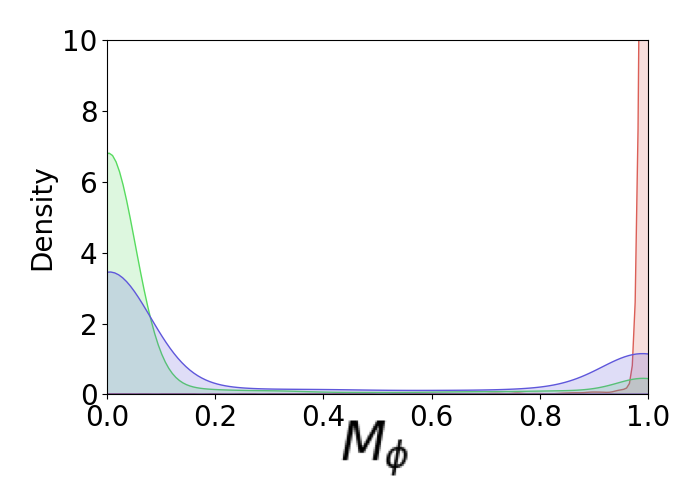} }}%
    \subfloat{{\includegraphics[width=0.48\columnwidth]{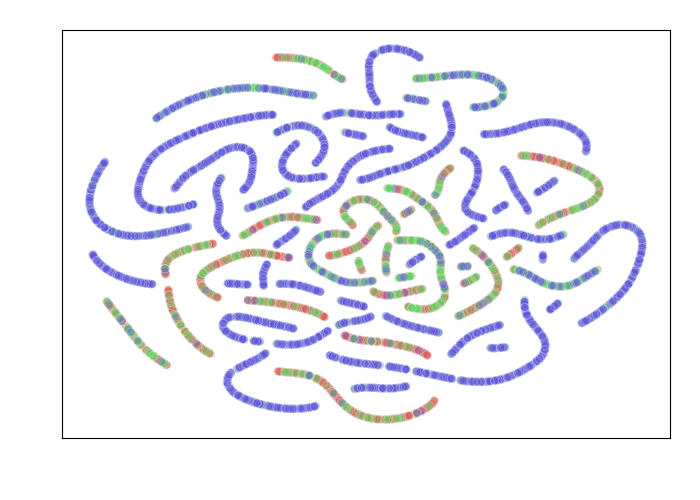} }}%
    \caption{$M_{\phi}$ densities and t-SNE of deviations for \texttt{hopper-medium-expert} (top row) and \texttt{Antmaze-large-diverse} (bottom row). Density plots are clipped at 10.0 as density for $\mathcal{D}$ is large. 10 actions are sampled from $\mathcal{D}_{\text{uni}}$ and $\mathcal{D}_{\text{perm}}$ each, per state.}
    \label{fig: morse analysis}
\end{figure}

\paragraph{Morse Network Analysis}We analyze how well the Morse network can distinguish between dataset tuples and samples from $\mathcal{D}_{\text{perm}}$, permutations of dataset actions, and $\mathcal{D}_{\text{uni}}$. We plot both certainty ($M_{\phi}$) density and t-SNEs \cite{RN591} in Figure~\ref{fig: morse analysis} which show that the unsupervised Morse network is effective in distinguishing between $\mathcal{D}_{\text{perm}}$ and $\mathcal{D}_{\text{uni}}$ and assigning high certainty to dataset tuples.

\begin{figure}[h]
    \centering
    \includegraphics[width=\columnwidth]{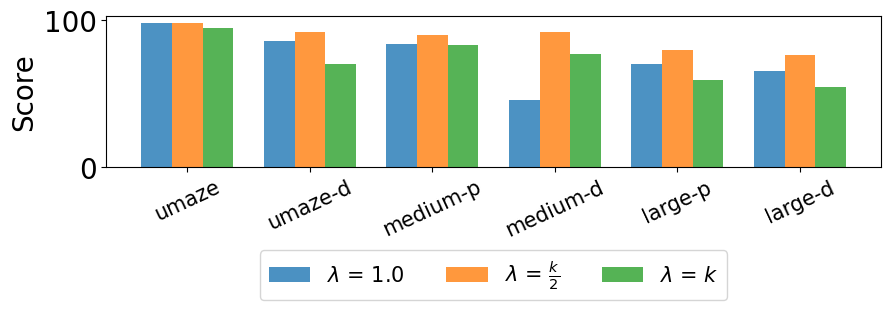}
    \caption{Ablations of $\lambda$ on Antmaze datasets. Recall $k = \dim(\mathcal{A})$.}
    \label{fig: antmaze lambda ablations}
\end{figure}

\begin{figure}[h]
    \centering
    \subfloat[\centering \texttt{hopper-medium}]{{\includegraphics[width=0.5\columnwidth]{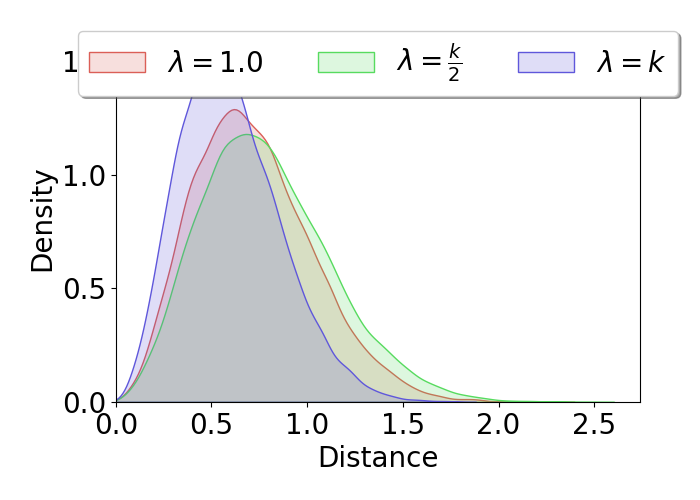} }}%
    \subfloat[\centering \texttt{amaze-large-play}]{{\includegraphics[width=0.5\columnwidth]{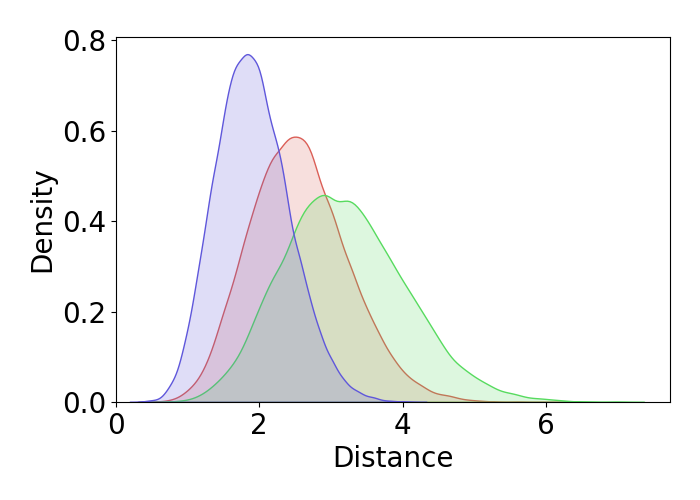} }}%
    \caption{Histograms of deviation from dataset actions.}
    \label{fig: tau deviations}
\end{figure}

\paragraph{Ablating kernel scale}We examine sensitivity to the kernel scale $\lambda$. Recall that $k = \dim(\mathcal{A})$. We see in Figure~\ref{fig: antmaze lambda ablations} that the scale $\lambda = \frac{k}{2}$ is a performance sweet-spot on the challenging Antmaze tasks. We further illustrate this by plotting policy deviations from dataset actions in Figure~\ref{fig: tau deviations}. The scale $\lambda = 1.0$ is potentially too lax a behavioral constraint, while $\lambda = k$ is too strong, resulting in performance reduction. However, performance on all scales remains strong and compares well with most prior algorithms. Performance may be further improved by tuning $\lambda$, possibly with separate scales for each input dimension.

\begin{figure}[h!]
    \centering
    \includegraphics[width=\columnwidth]{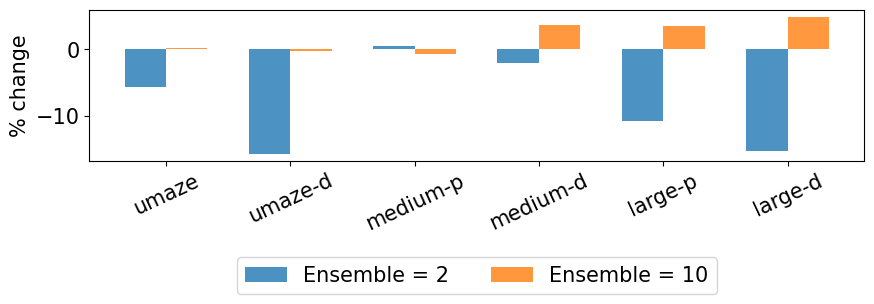}
    \caption{\% change in Antmaze scores without CDQ for critic ensembles consisting of 2 and 10 $Q$ functions.}
    \label{fig: antmaze cdq ablations}
\end{figure}

\paragraph{Independent or Shared Targets?}Standard TD3 employs Clipped Double $Q$-learning (CDQ) \cite{RN716,RN714} to prevent value overestimation. On tasks with sparse rewards, this may be too conservative \cite{RN883}. MSG \cite{RN771} uses large ensembles of fully independent $Q$ functions to learn offline. We examine how independent double $Q$ functions perform compared to the standard CDQ setup in Antmaze with 2 and 10 critics. The results in Figure~\ref{fig: antmaze cdq ablations} show that disabling CDQ with 2 critics is consistently detrimental to performance. Using a larger 10-critic ensemble leads to moderate improvements. This suggests that combining policy regularization with an efficient, independent ensemble could bring further performance benefits with minimal changes to the algorithm. 

\section{Discussion}

\paragraph{Morse Network}In \cite{RN870}, deeper architectures are required even when training on simple datasets. This rings true for our application of Morse networks in this work, with low-capacity networks performing poorly. Training the Morse network for each locomotion and Antmaze dataset typically takes 10 minutes for 100\,000 gradient steps using a batch size of 1\,024. When training the policy, using the Morse network increases training time by approximately 15\%.

\paragraph{Optimal Datasets}On Gym Locomotion tasks TD3-BST performance is comparable to newer methods, all of which rarely outperform older baselines. This can be attributed to a significant proportion of high-return-yielding trajectories that are easier to improve.

\section{Conclusion}

In this paper, we introduce TD3-BST, an algorithm that uses an uncertainty model to dynamically adjust the strength of regularization. Dynamic weighting allows the policy to maximize reward around individual dataset modes. Our algorithm compares well against prior methods on Gym Locomotion tasks and achieves the best scores on the more challenging Antmaze tasks, demonstrating strong performance when learning from suboptimal data. 

In addition, our experiments show that combining our policy regularization with an ensemble-based source of uncertainty can improve performance. Future work can explore how to apply alternative uncertainty measures and how best to combine multiple sources of uncertainty.


\bibliographystyle{named}
\bibliography{mybib}

\begin{thebibliography}{}

\bibitem[\protect\citeauthoryear{Amari}{2016}]{RN874}
Shun-ichi Amari.
\newblock {\em Information geometry and its applications}, volume 194.
\newblock Springer, 2016.

\bibitem[\protect\citeauthoryear{An \bgroup \em et al.\egroup }{2021}]{RN770}
Gaon An, Seungyong Moon, Jang-Hyun Kim, and Hyun~Oh Song.
\newblock Uncertainty-based offline reinforcement learning with diversified {Q}-ensemble.
\newblock {\em Advances in neural information processing systems}, 34:7436--7447, 2021.

\bibitem[\protect\citeauthoryear{Basu and Prasad}{2020}]{RN872}
Somnath Basu and Sachchidanand Prasad.
\newblock A connection between cut locus, thom space and morse-bott functions.
\newblock {\em arXiv preprint arXiv:2011.02972}, 2020.

\bibitem[\protect\citeauthoryear{Brandfonbrener \bgroup \em et al.\egroup }{2021}]{RN882}
David Brandfonbrener, Will Whitney, Rajesh Ranganath, and Joan Bruna.
\newblock Offline {RL} without off-policy evaluation.
\newblock {\em Advances in neural information processing systems}, 34:4933--4946, 2021.

\bibitem[\protect\citeauthoryear{Dherin \bgroup \em et al.\egroup }{2023}]{RN870}
Benoit Dherin, Huiyi Hu, Jie Ren, Michael~W Dusenberry, and Balaji Lakshminarayanan.
\newblock Morse neural networks for uncertainty quantification.
\newblock {\em arXiv preprint arXiv:2307.00667}, 2023.

\bibitem[\protect\citeauthoryear{Emerson \bgroup \em et al.\egroup }{2023}]{RN855}
Harry Emerson, Matthew Guy, and Ryan McConville.
\newblock Offline reinforcement learning for safer blood glucose control in people with type 1 diabetes.
\newblock {\em Journal of Biomedical Informatics}, 142:104376, 2023.

\bibitem[\protect\citeauthoryear{Florence \bgroup \em et al.\egroup }{2022}]{RN885}
Pete Florence, Corey Lynch, Andy Zeng, Oscar~A Ramirez, Ayzaan Wahid, Laura Downs, Adrian Wong, Johnny Lee, Igor Mordatch, and Jonathan Tompson.
\newblock Implicit behavioral cloning.
\newblock In {\em Conference on Robot Learning}, pages 158--168. PMLR, 2022.

\bibitem[\protect\citeauthoryear{Fu \bgroup \em et al.\egroup }{2020}]{RN731}
Justin Fu, Aviral Kumar, Ofir Nachum, George Tucker, and Sergey Levine.
\newblock {D4RL}: Datasets for deep data-driven reinforcement learning.
\newblock {\em arXiv preprint arXiv:2004.07219}, 2020.

\bibitem[\protect\citeauthoryear{Fu \bgroup \em et al.\egroup }{2022}]{RN813}
Yuwei Fu, Di~Wu, and Benoit Boulet.
\newblock A closer look at offline {RL} agents.
\newblock {\em Advances in Neural Information Processing Systems}, 35:8591--8604, 2022.

\bibitem[\protect\citeauthoryear{Fujimoto and Gu}{2021}]{RN719}
Scott Fujimoto and Shixiang~Shane Gu.
\newblock A minimalist approach to offline reinforcement learning.
\newblock {\em Advances in neural information processing systems}, 34:20132--20145, 2021.

\bibitem[\protect\citeauthoryear{Fujimoto \bgroup \em et al.\egroup }{2018}]{RN714}
Scott Fujimoto, Herke Hoof, and David Meger.
\newblock Addressing function approximation error in actor-critic methods.
\newblock In {\em International conference on machine learning}, pages 1587--1596. PMLR, 2018.

\bibitem[\protect\citeauthoryear{Fujimoto \bgroup \em et al.\egroup }{2019}]{RN684}
Scott Fujimoto, David Meger, and Doina Precup.
\newblock Off-policy deep reinforcement learning without exploration.
\newblock In {\em International conference on machine learning}, pages 2052--2062. PMLR, 2019.

\bibitem[\protect\citeauthoryear{Ghasemipour \bgroup \em et al.\egroup }{2022}]{RN771}
Kamyar Ghasemipour, Shixiang~Shane Gu, and Ofir Nachum.
\newblock Why so pessimistic? estimating uncertainties for offline {RL} through ensembles, and why their independence matters.
\newblock {\em Advances in Neural Information Processing Systems}, 35:18267--18281, 2022.

\bibitem[\protect\citeauthoryear{Goodfellow \bgroup \em et al.\egroup }{2014}]{RN864}
Ian~J Goodfellow, Jonathon Shlens, and Christian Szegedy.
\newblock Explaining and harnessing adversarial examples.
\newblock {\em arXiv preprint arXiv:1412.6572}, 2014.

\bibitem[\protect\citeauthoryear{Goodfellow \bgroup \em et al.\egroup }{2016}]{RN884}
Ian Goodfellow, Yoshua Bengio, and Aaron Courville.
\newblock {\em Deep learning}.
\newblock MIT press, 2016.

\bibitem[\protect\citeauthoryear{Gulcehre \bgroup \em et al.\egroup }{2020}]{RN841}
Caglar Gulcehre, Sergio~Gómez Colmenarejo, Jakub Sygnowski, Thomas Paine, Konrad Zolna, Yutian Chen, Matthew Hoffman, Razvan Pascanu, and Nando de~Freitas.
\newblock Addressing extrapolation error in deep offline reinforcement learning.
\newblock 2020.

\bibitem[\protect\citeauthoryear{Hasselt}{2010}]{RN716}
Hado Hasselt.
\newblock Double {Q}-learning.
\newblock {\em Advances in neural information processing systems}, 23, 2010.

\bibitem[\protect\citeauthoryear{Kondratyuk \bgroup \em et al.\egroup }{2020}]{RN860}
Dan Kondratyuk, Mingxing Tan, Matthew Brown, and Boqing Gong.
\newblock When ensembling smaller models is more efficient than single large models.
\newblock {\em arXiv preprint arXiv:2005.00570}, 2020.

\bibitem[\protect\citeauthoryear{Kostrikov \bgroup \em et al.\egroup }{2021a}]{RN703}
Ilya Kostrikov, Rob Fergus, Jonathan Tompson, and Ofir Nachum.
\newblock Offline reinforcement learning with {F}isher divergence critic regularization.
\newblock In {\em International Conference on Machine Learning}, pages 5774--5783. PMLR, 2021.

\bibitem[\protect\citeauthoryear{Kostrikov \bgroup \em et al.\egroup }{2021b}]{RN711}
Ilya Kostrikov, Ashvin Nair, and Sergey Levine.
\newblock Offline reinforcement learning with implicit {Q}-learning.
\newblock {\em arXiv preprint arXiv:2110.06169}, 2021.

\bibitem[\protect\citeauthoryear{Kumar \bgroup \em et al.\egroup }{2019}]{RN697}
Aviral Kumar, Justin Fu, Matthew Soh, George Tucker, and Sergey Levine.
\newblock Stabilizing off-policy {Q}-learning via bootstrapping error reduction.
\newblock {\em Advances in Neural Information Processing Systems}, 32, 2019.

\bibitem[\protect\citeauthoryear{Kumar \bgroup \em et al.\egroup }{2020}]{RN698}
Aviral Kumar, Aurick Zhou, George Tucker, and Sergey Levine.
\newblock Conservative {Q}-learning for offline reinforcement learning.
\newblock {\em Advances in Neural Information Processing Systems}, 33:1179--1191, 2020.

\bibitem[\protect\citeauthoryear{Lakshminarayanan \bgroup \em et al.\egroup }{2017}]{RN868}
Balaji Lakshminarayanan, Alexander Pritzel, and Charles Blundell.
\newblock Simple and scalable predictive uncertainty estimation using deep ensembles.
\newblock {\em Advances in neural information processing systems}, 30, 2017.

\bibitem[\protect\citeauthoryear{Lange \bgroup \em et al.\egroup }{2012}]{RN695}
Sascha Lange, Thomas Gabel, and Martin Riedmiller.
\newblock {\em Batch reinforcement learning}, pages 45--73.
\newblock Springer, 2012.

\bibitem[\protect\citeauthoryear{Lee \bgroup \em et al.\egroup }{2018}]{RN869}
Kimin Lee, Kibok Lee, Honglak Lee, and Jinwoo Shin.
\newblock A simple unified framework for detecting out-of-distribution samples and adversarial attacks.
\newblock {\em Advances in neural information processing systems}, 31, 2018.

\bibitem[\protect\citeauthoryear{Lei \bgroup \em et al.\egroup }{2023}]{RN880}
Kun Lei, Zhengmao He, Chenhao Lu, Kaizhe Hu, Yang Gao, and Huazhe Xu.
\newblock Uni-o4: Unifying online and offline deep reinforcement learning with multi-step on-policy optimization.
\newblock {\em arXiv preprint arXiv:2311.03351}, 2023.

\bibitem[\protect\citeauthoryear{Li \bgroup \em et al.\egroup }{2022}]{RN844}
Jianxiong Li, Xianyuan Zhan, Haoran Xu, Xiangyu Zhu, Jingjing Liu, and Ya-Qin Zhang.
\newblock When data geometry meets deep function: Generalizing offline reinforcement learning.
\newblock In {\em The Eleventh International Conference on Learning Representations}, 2022.

\bibitem[\protect\citeauthoryear{Li \bgroup \em et al.\egroup }{2023}]{RN876}
Jiachen Li, Edwin Zhang, Ming Yin, Qinxun Bai, Yu-Xiang Wang, and William~Yang Wang.
\newblock Offline reinforcement learning with closed-form policy improvement operators.
\newblock In {\em International Conference on Machine Learning}, pages 20485--20528. PMLR, 2023.

\bibitem[\protect\citeauthoryear{Mirowski \bgroup \em et al.\egroup }{2018}]{RN839}
Piotr Mirowski, Matt Grimes, Mateusz Malinowski, Karl~Moritz Hermann, Keith Anderson, Denis Teplyashin, Karen Simonyan, Andrew Zisserman, and Raia Hadsell.
\newblock Learning to navigate in cities without a map.
\newblock {\em Advances in neural information processing systems}, 31, 2018.

\bibitem[\protect\citeauthoryear{Moskovitz \bgroup \em et al.\egroup }{2021}]{RN883}
Ted Moskovitz, Jack Parker-Holder, Aldo Pacchiano, Michael Arbel, and Michael Jordan.
\newblock Tactical optimism and pessimism for deep reinforcement learning.
\newblock {\em Advances in Neural Information Processing Systems}, 34:12849--12863, 2021.

\bibitem[\protect\citeauthoryear{Nair \bgroup \em et al.\egroup }{2020}]{RN706}
Ashvin Nair, Abhishek Gupta, Murtaza Dalal, and Sergey Levine.
\newblock {AWAC}: Accelerating online reinforcement learning with offline datasets.
\newblock {\em arXiv preprint arXiv:2006.09359}, 2020.

\bibitem[\protect\citeauthoryear{Nguyen \bgroup \em et al.\egroup }{2015}]{RN863}
Anh Nguyen, Jason Yosinski, and Jeff Clune.
\newblock Deep neural networks are easily fooled: High confidence predictions for unrecognizable images.
\newblock In {\em Proceedings of the IEEE conference on computer vision and pattern recognition}, pages 427--436, 2015.

\bibitem[\protect\citeauthoryear{Nikulin \bgroup \em et al.\egroup }{2023}]{RN847}
Alexander Nikulin, Vladislav Kurenkov, Denis Tarasov, and Sergey Kolesnikov.
\newblock Anti-exploration by random network distillation.
\newblock {\em arXiv preprint arXiv:2301.13616}, 2023.

\bibitem[\protect\citeauthoryear{Peng \bgroup \em et al.\egroup }{2019}]{RN705}
Xue~Bin Peng, Aviral Kumar, Grace Zhang, and Sergey Levine.
\newblock Advantage-weighted regression: Simple and scalable off-policy reinforcement learning.
\newblock {\em arXiv preprint arXiv:1910.00177}, 2019.

\bibitem[\protect\citeauthoryear{Rezaeifar \bgroup \em et al.\egroup }{2022}]{RN723}
Shideh Rezaeifar, Robert Dadashi, Nino Vieillard, Léonard Hussenot, Olivier Bachem, Olivier Pietquin, and Matthieu Geist.
\newblock Offline reinforcement learning as anti-exploration.
\newblock In {\em Proceedings of the AAAI Conference on Artificial Intelligence}, volume~36, pages 8106--8114, 2022.

\bibitem[\protect\citeauthoryear{Rummery and Niranjan}{1994}]{RN886}
Gavin~A Rummery and Mahesan Niranjan.
\newblock {\em On-line Q-learning using connectionist systems}, volume~37.
\newblock University of Cambridge, Department of Engineering Cambridge, UK, 1994.

\bibitem[\protect\citeauthoryear{Singh \bgroup \em et al.\egroup }{2022}]{RN754}
Anikait Singh, Aviral Kumar, Quan Vuong, Yevgen Chebotar, and Sergey Levine.
\newblock Offline {RL} with realistic datasets: Heteroskedasticity and support constraints.
\newblock {\em arXiv preprint arXiv:2211.01052}, 2022.

\bibitem[\protect\citeauthoryear{Sutton and Barto}{2018}]{RN679}
Richard~S Sutton and Andrew~G Barto.
\newblock {\em Reinforcement learning: An introduction}.
\newblock MIT press, 2018.

\bibitem[\protect\citeauthoryear{Tang and Wiens}{2021}]{RN854}
Shengpu Tang and Jenna Wiens.
\newblock Model selection for offline reinforcement learning: Practical considerations for healthcare settings.
\newblock In {\em Machine Learning for Healthcare Conference}, pages 2--35. PMLR, 2021.

\bibitem[\protect\citeauthoryear{Tarasov \bgroup \em et al.\egroup }{2022}]{RN820}
Denis Tarasov, Alexander Nikulin, Dmitry Akimov, Vladislav Kurenkov, and Sergey Kolesnikov.
\newblock {CORL}: Research-oriented deep offline reinforcement learning library.
\newblock {\em arXiv preprint arXiv:2210.07105}, 2022.

\bibitem[\protect\citeauthoryear{Tarasov \bgroup \em et al.\egroup }{2023}]{RN842}
Denis Tarasov, Vladislav Kurenkov, Alexander Nikulin, and Sergey Kolesnikov.
\newblock Revisiting the minimalist approach to offline reinforcement learning.
\newblock {\em arXiv preprint arXiv:2305.09836}, 2023.

\bibitem[\protect\citeauthoryear{Van~der Maaten and Hinton}{2008}]{RN591}
Laurens Van~der Maaten and Geoffrey Hinton.
\newblock Visualizing data using t-{SNE}.
\newblock {\em Journal of machine learning research}, 9(11), 2008.

\bibitem[\protect\citeauthoryear{Williams and Rasmussen}{2006}]{RN887}
Christopher~KI Williams and Carl~Edward Rasmussen.
\newblock {\em Gaussian processes for machine learning}, volume~2.
\newblock MIT press Cambridge, MA, 2006.

\bibitem[\protect\citeauthoryear{Wu \bgroup \em et al.\egroup }{2019}]{RN699}
Yifan Wu, George Tucker, and Ofir Nachum.
\newblock Behavior regularized offline reinforcement learning.
\newblock {\em arXiv preprint arXiv:1911.11361}, 2019.

\bibitem[\protect\citeauthoryear{Xu \bgroup \em et al.\egroup }{2023}]{RN851}
Haoran Xu, Li~Jiang, Jianxiong Li, Zhuoran Yang, Zhaoran Wang, Victor Wai~Kin Chan, and Xianyuan Zhan.
\newblock Offline {RL} with no {OOD} actions: In-sample learning via implicit value regularization.
\newblock {\em arXiv preprint arXiv:2303.15810}, 2023.

\bibitem[\protect\citeauthoryear{Yu \bgroup \em et al.\egroup }{2021}]{RN840}
Chao Yu, Jiming Liu, Shamim Nemati, and Guosheng Yin.
\newblock Reinforcement learning in healthcare: A survey.
\newblock {\em ACM Computing Surveys (CSUR)}, 55(1):1--36, 2021.

\bibitem[\protect\citeauthoryear{Zhang and Jiang}{2021}]{RN852}
Siyuan Zhang and Nan Jiang.
\newblock Towards hyperparameter-free policy selection for offline reinforcement learning.
\newblock {\em Advances in Neural Information Processing Systems}, 34:12864--12875, 2021.

\bibitem[\protect\citeauthoryear{Zhang \bgroup \em et al.\egroup }{2023}]{RN873}
Hongchang Zhang, Jianzhun Shao, Shuncheng He, Yuhang Jiang, and Xiangyang Ji.
\newblock {DARL}: distance-aware uncertainty estimation for offline reinforcement learning.
\newblock In {\em Proceedings of the AAAI Conference on Artificial Intelligence}, volume~37, pages 11210--11218, 2023.

\bibitem[\protect\citeauthoryear{Zhu \bgroup \em et al.\egroup }{2022}]{RN843}
Deyao Zhu, Li~Erran Li, and Mohamed Elhoseiny.
\newblock Value memory graph: A graph-structured world model for offline reinforcement learning.
\newblock {\em arXiv preprint arXiv:2206.04384}, 2022.

\bibitem[\protect\citeauthoryear{Zhu \bgroup \em et al.\egroup }{2023}]{RN856}
Taiyu Zhu, Kezhi Li, and Pantelis Georgiou.
\newblock Offline deep reinforcement learning and off-policy evaluation for personalized basal insulin control in type 1 diabetes.
\newblock {\em IEEE Journal of Biomedical and Health Informatics}, 2023.

\end{thebibliography}

\end{document}